# A Natural Language Processing Framework for Hotel Recommendation based on user's text reviews.

**Lavrentia Aravani, Emmanuel Pintelas, Christos Pierrakeas, Panagiotis Pintelas**

---

**L. ARAVANI**
Management Science and Technology Department, University of Patras, Patras, Greece
**E. PINTELAS**
Department of Mathematics, University of Patras, Patras, Greece
**C. PIERRAKEAS**
Management Science and Technology Department, University of Patras, Patras, Greece
**P. PINTELAS**
Department of Mathematics, University of Patras, Patras, Greece

## Abstract

Recently, the application of Artificial Intelligence algorithms in hotel recommendation systems has become an increasingly popular topic. One such method that has proven to be effective in this field is Deep Learning, especially Natural Language processing models, which are able to extract semantic knowledge from user's text reviews to create more efficient recommendation systems. This can lead to the development of intelligent models that can classify a user's preferences and emotions based on their feedback in the form of text reviews about their hotel stay experience. In this study, we propose a Natural Language Processing framework that utilizes customer text reviews to provide personalized recommendations for the most appropriate hotel based on their preferences. The framework is based on Bidirectional Encoder Representations from Transformers (BERT) and a fine-tuning/validation pipeline that categorizes customer hotel review texts into "Bad," "Good," or "Excellent" recommended hotels. Our findings indicate that the hotel recommendation system we propose can significantly enhance the user experience of booking accommodations by providing personalized recommendations based on user preferences and previous booking history.

## 1. Introduction

The growing significance of online reviews in shaping consumer choices within the dynamic hospitality industry cannot be overstated. As the digital age progresses, online platforms, particularly travel review sites, have emerged as crucial information hubs for individuals seeking lodging. The widespread use of social media has accelerated the dissemination of information by allowing individuals to easily share their opinions, experiences, and criticisms.

This study proposes a Natural Language Processing (NLP) framework designed to revolutionize the field of hotel recommendations. With an increasing number of travelers turning to online reviews for guidance in their decision-making, current platforms often fail to effectively filter and rank pertinent information. Prominent travel review websites, such as Booking.com, TripAdvisor.com, and Agoda.com, typically arrange reviews based on the date they were posted or the number of votes they have received. While this approach appears to provide some order, it unintentionally disregards more recent reviews, thereby diminishing their influence on users' decision-making.

To address this limitation, our research capitalizes on the wealth of data present in user-generated text reviews to propose a unified recommendation framework based on natural language processing techniques. Our objective is to overcome the constraints of existing review ranking systems through the use of artificial intelligence and machine learning. This paper presents a comprehensive method for extracting valuable insights from textual reviews, thereby enhancing the accuracy and applicability of hotel recommendations.

We analyze the state-of-the-art in text classification models in this context and draw inspiration from various recommender system approaches. The proposed framework integrates artificial intelligence (AI), machine learning classification, and sentiment analysis methods to create a reliable Decision Support System (DSS). This DSS, powered by natural language processing (NLP) algorithms, aims to comprehend and classify the sentiment and preferences conveyed in user reviews, ultimately assisting travelers in making well-informed decisions that align with their individual preferences.
The main contributions of our research work are summarized as follows:

- We propose a unified recommendation framework based on hotel text reviews which manages to achieve superior performance.
- We propose a training/validation pipeline which significantly boost the representational capabilities of BERT backbone in the hotel recommendation context.
- Finally, our hotel recommendation framework can potentially offer a number of benefits, including informed decision-making, personalization, customer empowerment, increased information accessibility, quality assurance, greater customer satisfaction, trust, and a feedback loop between customers and hotels.

The remainder of this paper is organized as follows. Section 2 presents the state-of-the-art works relative to Text Classification models. Section 3 presents a detailed discussion about the proposed framework while Section 4 presents our Experimental Results. Finally, Section 5 sketches our conclusive remarks and possible future work.

## 2. Related Work

Scholars have been interested in studies on hotel recommendation systems. Two classes can be distinguished from the studies: The hotel recommendation systems that use numerical rating data are included in the first class. The hotel recommendation systems that use text review data are included in the second class [9-15].

### 2.1 TfidfVectorizer

TfidfVectorizer is a text vectorization technique that is commonly used in natural language processing (NLP) to transform text documents into numerical feature vectors that can be used as input to machine learning models.

The TfidfVectorizer calculates the Term Frequency-Inverse Document Frequency (TF-IDF) values for each term in the document corpus. TF-IDF is a statistical measure that reflects the importance of a word in a document relative to its importance in the entire corpus of documents. To assess the link between each word in the collection of documents, the TF-IDF has been extensively utilized in the domains of text mining and information retrieval. They are specifically used to determine search ranking, compute comparable degrees among documents, extract core words (i.e., keywords) from papers, and other tasks.

Text preprocessing, information retrieval, document clustering, and text classification are among the uses of TF-IDF. The TfidfVectorizer first converts each text document into a bag of words, which is a simple representation of the text where the order of the words is ignored and only the frequency of each word is considered. Then, for each word in the bag of words, the TfidfVectorizer calculates the TF-IDF value, which is the product of the term frequency (TF) and inverse document frequency (IDF) values.

The term frequency (TF) is simply the frequency of a term in a document, while the inverse document frequency (IDF) measures how common or rare a term is across all documents in the corpus. The IDF value is calculated as the logarithm of the total number of documents in the corpus divided by the number of documents containing the term.

The resulting TF-IDF values for each term in each document form the feature vectors that can be used as input to machine learning models. The TfidfVectorizer also provides options for text preprocessing, such as tokenization and stop word removal, to further improve the quality of the feature vectors. Its benefits include feature selection, memory efficiency through sparse representation, and adaptability to a range of natural language processing applications.

In summary, TfidfVectorizer is a widely used text vectorization technique that calculates TF-IDF values for each term in a document corpus, resulting in numerical feature vectors that can be used as input to machine learning models in NLP tasks such as text classification and clustering.

### 2.2. RNNs

Recurrent neural networks (RNNs) are a type of neural network used to process sequential data, such as time series or natural language. RNNs differ from traditional deep neural networks in that they have the ability to retain information from previous steps and can use it to produce the current output. This means that it is affected by all previous inputs. These deep learning algorithms are built into well-known programs like Siri, voice search, and Google Translate. That is, they are used for syntactic or temporal problems such as language translation, natural language processing (nlp), speech recognition, and image captioning. Recurrent neural networks also face some limitations. More specifically, they are unable to incorporate future data, which means that they only consider information from the past and not the future when forecasting. Although they have some limitations, their ability to leverage prior information makes them valuable in many fields.

To help us understand RNNs, we can use a familiar expression like "feeling under the weather." This phrase is an English idiomatic expression and means that someone feels sick or unwell. This expression only makes sense if the words are placed in the correct order. RNNs must therefore take into account the position of each word and use this knowledge to predict the next word in the sequence.

RNNs are also distinguished by the fact that their parameters are common to all layers of the network. This means that recurrent networks share the same weight parameter across all layers, unlike feedforward networks that have separate weights for each node. Although the weights are modified through processes such as gradient descent and backpropagation to support reinforcement learning, RNNs often face problems with exploding or vanishing gradients.

The size of the slope determines these problems. When the slope becomes too small, the changes in the weight parameters become negligible, and the network stops learning. When the gradient becomes too large, it can explode, making the model unstable with the weights becoming too large and represented as NaN. Reducing the number of hidden layers can help address these issues.

The key property of RNNs is their ability to retain and transfer information over time. This is achieved through recurrent connections between hidden layers, which allow the network to "remember" previous inputs. This memory makes RNNs suitable for tasks where the input is a sequence of data, such as speech recognition, language translation, and text analysis. In the architecture of an RNN, the input is fed to the network step by step, and at each time step, the current input is combined with the hidden state from the previous step to produce a new hidden state. This new hidden state is passed on to the next step, allowing the network to maintain a persistent memory of previous inputs.

In general, RNNs are ideal for tasks where the order and temporal dimension of the data are important, and they can adjust their predictions based on the sequence of inputs.

### 2.3 NLP Transformers

NLP Transformers are a type of neural network architecture that has revolutionized the field of natural language processing (NLP) by enabling state-of-the-art performance on a wide range of tasks.

Transformers are based on a self-attention mechanism, which allows the network to weigh different parts of the input sequence differently when making predictions. This contrasts with traditional recurrent neural networks (RNNs), which process input sequences one element at a time and maintain a hidden state that summarizes the sequence so far.

The Transformer architecture was introduced in the paper "Attention Is All You Need" by Vaswani et al. (2017) [16] and has since become the de facto standard for NLP tasks such as language modeling, machine translation, and text classification.

In a transformer network, the input sequence is first embedded into a high-dimensional vector space. The network then applies multiple layers of self-attention and feedforward neural networks to the embedded sequence, with residual connections and layer normalization between layers to help with training stability.

One of the key advantages of transformer networks is their ability to model long-range dependencies in input sequences. This is achieved through the self-attention mechanism, which allows the network to attend to any part of the input sequence regardless of its position.

Another advantage of transformers is their ability to be pre-trained on large amounts of unlabeled text data using techniques such as masked language modeling and next sentence prediction. This pre-training can then be fine-tuned on specific downstream NLP tasks with relatively small amounts of labeled data, leading to state-of-the-art performance on these tasks.

Overall, NLP transformers have significantly advanced the state of the art in natural language processing, enabling more accurate and efficient models for a wide range of applications.

## 3 Proposed Framework

In this paper, we put forth a new technique that employs text reviews from customers and develops a potent Decision Support System (DSS) to aid hotel guests in selecting a hotel that aligns with their financial and other preferences. This methodology incorporates machine learning (ML), artificial intelligence (AI), and natural language processing (NLP) techniques.

In the suggested approach, a hotel feature matrix is extracted using opinion-based sentiment analysis and saved in a database. To ascertain how people feel about hotel amenities, our method combines lexical analysis with syntax analysis and semantic analysis.

The NLP text classification framework proposed for this task is based on the pre-trained BERT architecture, which enables the extraction of features from raw text data and their subsequent input into Fully-Connected (FC) neural networks. This configuration enables the classification of client hotel reviews as "Bad," "Good," or "Excellent," and recommends hotels accordingly.

In order to deliver suggestions based on dynamic data discovered concurrently with active user queries, the recommender system must be developed in a way that will employ dynamic auto updated data containing the visual views, votes, and reviews online from the external websites. Web cookies, customer browsing activity, and comments on the new recommended goods will all be used to increase the flexibility of recommender services.

The full implementation code for this project can be accessed via the following link: <https://github.com/EmmanuelPintelas/BERT-V-Training-Tool-Box-for-Text-Classification>.

More specifically, the proposed NLP text classification framework is based on the pre-trained BERT architecture in order to extract features from raw text data, feeding them into Fully-Connected (FC) neural networks in order to classify the client's hotel review texts, into "Bad", "Good", or "Excellent" recommended hotel.

An example of a person's review text which was tagged as "Good" is presented below:

"Enjoyed our stay! We enjoyed our stay at the Century City Hyatt. We booked two adjoining rooms as we were traveling as a family with three kids. Clearly more of a business hotel than a family oriented hotel but for the most part no problems with kids. Read the listing by another member that they served free Starbucks coffee to Gold passport members. I am a Platinum Hyatt Gold passport member and a little disappointed I wasn't aware of that offer! We were pleased with the rooms - nice sized with a view of the pool. We liked the decks off of the rooms. Very friendly service and great location. The hotel is directly in the middle of the two places we wanted to go while there - Hollywood and Santa Monica. The one disappointment - We were rather offended when we were near the spa/fitness center on the pool side and decided to walk through to get to the mall across the street and were treated rudely by a spa employee. We could see the door to exit onto the street but she insisted we walk back through the spa, through the pool area, and through the hotel, then walk all around the block as no children were allowed in the spa. Our children were not loud toddlers - they are 7, 10, and 17! First of all, there was no sign on the spa/fitness door asking that children not enter. Secondly, the spa front desk is right next to the main entrance to the spa. Therefore I am convinced that by the time we had gotten to the desk the spa clients would not have been inconvenienced by us continuing through. When I said we were hotel guests at the Hyatt and trying to save some time crossing the street to the woman at the spa she said, We are not a part of the Hyatt. Yet the hotel lists it as their spa and the Hyatt door attendants also appear to work their door as well (same outfits, etc). I had planned to book a spa service during our stay but decided against it due to that one interaction. Beyond the spa staff, we were pleased with the hotel and would definitely stay there again."

An example of a person's review text which was tagged as "Bad" is presented below: "Disappointed We booked this hotel entirely due to the favourable reviews on this site. And while there were some nice things about it (I'll get to those), overall we were quite disappointed. The whole time I was there, I kept thinking, I can't wait to write a review of this place on TripAdvisor. The main problem was the tiny, cramped rooms (at least the one we had, it was a standard-type room with 2 queen beds). Everyone's face fell when we first walked in. It was claustrophobic. I have re-read many of the reviews now that I'm back, and even many of the favourable ones do mention the small rooms.The bathroom was equally cramped, with only space for a shower stall. This disappointed me further, as I am a bath person. Another disappointment was the lack of a fridge, something we always appreciate in a hotel room.Another disappointment was the size of the swimming pool (not much bigger than a bathtub) and the cold temperature of the water.The courtyard itself was nicely done up, in particular at night with the lights on the palm trees. Breakfast was fairly uninteresting (continental style danishes etc.), but I'm not really complaining here -- I'd rather have a simple breakfast than none at all. Cookies offered every afternoon at 4:00 were appreciated by my kids. Another thing I did like was the hair products, moisturizer and soaps. They were lovely.Next time we stay in San Fran, we will definitely try somewhere else."

An example of a person's review text which was tagged as "Excellent" is presented below: "Wow, what charm! As a Travel Agent, I've stayed at quite a few hotel, but this is the only Historic hotel so far... I loved it! Had to go back for a personal stay. The decor is beautiful, the lobby furniture fits the time period is still comfy. The city view rooms are great - love the little balconies. Great breakfast, nice people, great location - The Seattle Underground Tours is a 1/2 block away. I've aready sent my folk there for a stay have told others."

Moreover, we developed a Validation (V) algorithm in order to efficiently train the NLP model, presented in Algorithm 1. During training we totally froze (no updates in network weights) the first 150 layers of BERT backbone, while we unfroze (let the network update its weights) its last 50 layers. The first layers remain frozen in order to avoid overfitting and catastrophic forgetting of pre-trained network's weights, while the last layers will fine-tune with respect to our hotel text reviewing classification task. The proposed text classification framework (BERT-V) is presented in Fig. 1, Fig. 2, and Algorithm 1.

The proposed validation algorithm mainly updates the learning rate of the training procedure with respect to the validation accuracy. Also, if no further improvement is performed, the model weights are switched back into the ones who managed to bring the highest latest score. This is performed in order to avoid local optimum solutions by finding the optimum network weights parameters which maximizes the final performance score.

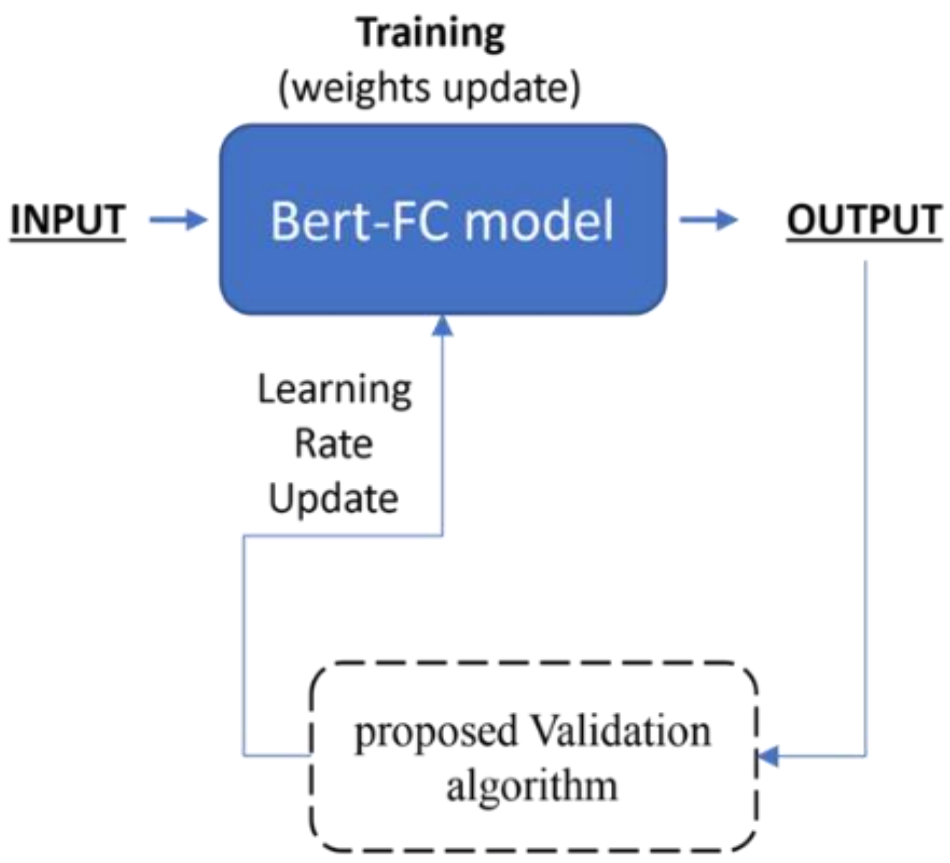

Figure 1 Abstract presentation of proposed text classification framework.

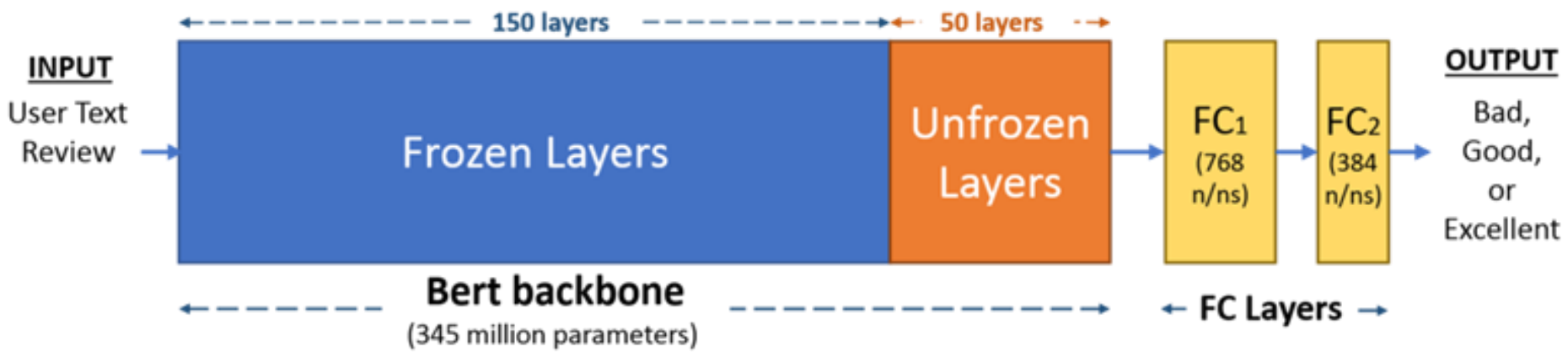

Figure 2 Architecture of BERT-FC model

**Algorithm 1.** Pseudocode of proposed Validation algorithm.

```
global cnt, best_sc, learning_rate_init = 0, 0, 1e-4
while Training == True:
        Input: Text
           pred = BERT_FC.predict (Text)
           sc = score_metric (pred, y)
           if sc > best_sc:
                  cnt = 0
                  best_sc = sc
                  best_weigths = BERT_FC.save()
           else:
                  cnt += 1
                  if cnt == 1:
                         learning_rate = learning_rate_init / 2
                  elif cnt == 2:
                         learning_rate = learning_rate_init / 10
                  elif cnt == 3:
                         learning_rate = learning_rate_init / 100
                  elif cnt > 3 and cnt <= 5:
                         learning_rate = rand_select ([1e-4, 1e-5,
                                         1e-6, 1e-7, 5e-8])
                  elif cnt == 6:
                         BERT_FC.load(best_weigths)
                         learning_rate = learning_rate_init
                         cnt = 0
        Output learning_rate
```

## 4. Experimental Results

In Table 1, we present the performance results of the proposed framework comparing to other state of the art text classification approaches, applied into our hotel text review classification problem. Based on our experimental results, the proposed framework outperformed all other approaches, revealing the superiority of our methodology.

We also performed a word-cloud analysis which allows as to identify and interpret the top-most frequent and important words with respect to the labels of our classification task ("Bad", "Good", or "Excellent") as presented in Figs. 3-5.

The full implementation code, alongside the dataset used in our experimental simulations can be found into the following link:
https://github.com/EmmanuelPintelas/BERT-V-Training-Tool-Box-for-Text-Classification

Table 1. Performance results of proposed framework comparing to other state of the art text classification approaches.

| | |
|---|---|
| TfidfVectorizer | 65.87 |
| RNNs | 61.12 |
| Transformer | 71.01 |
| BERT | 74.15 |
| BERT-V | 74.58 |

**Fig. 3** Top 50 Words: "Bad" tagged hotels.

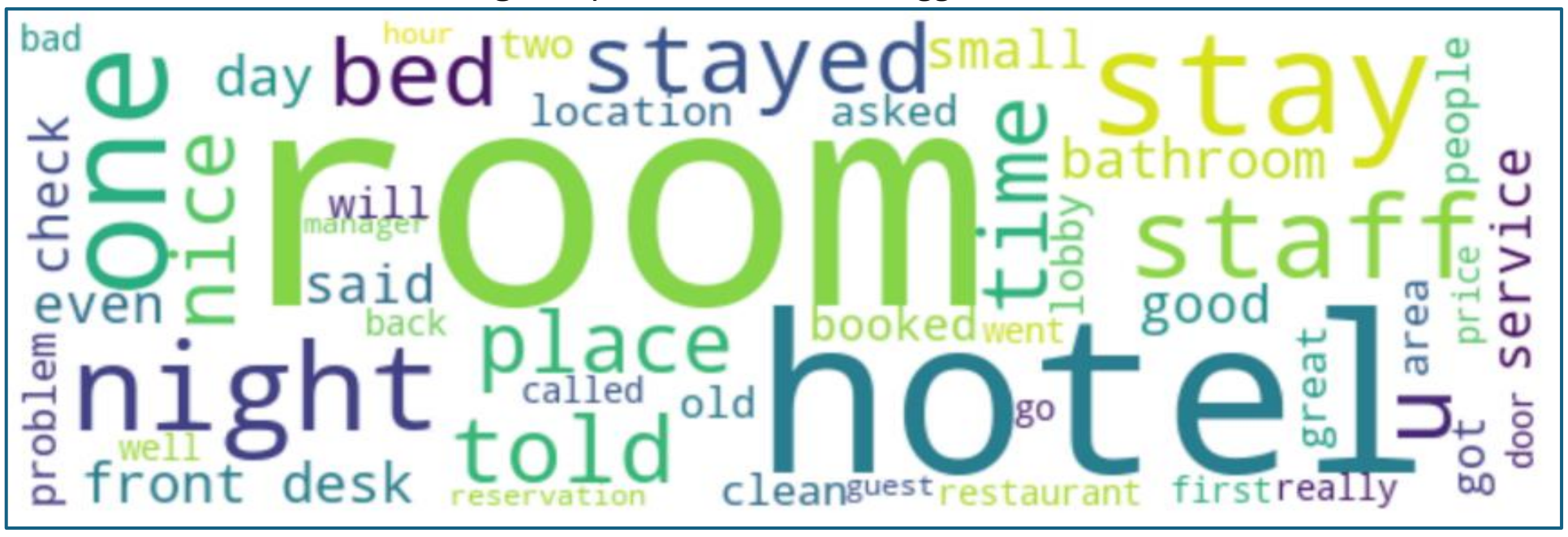

We can identify some words which indicate the hotels as a "bad" choice, such as "problem", "small", "old".

**Fig. 4** Top 50 Words: "Good" tagged hotels

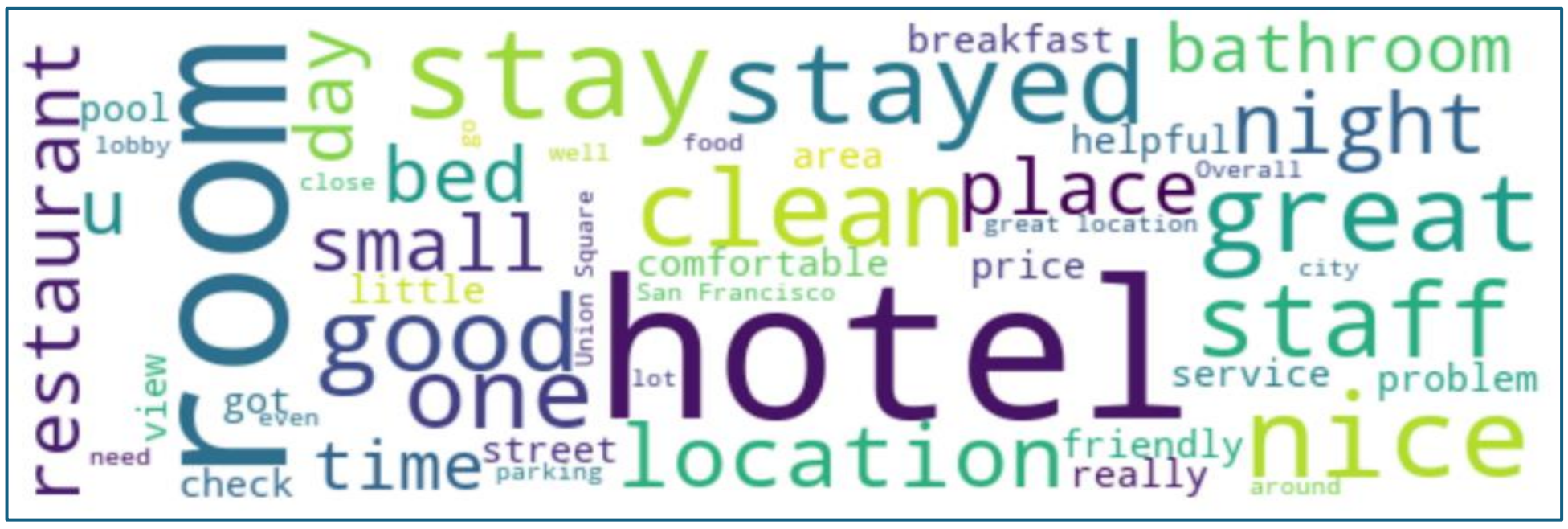

We can identify some words which indicate the hotels as a “good” choice, such as “clean”, “great”, “nice”, “friendly”.

**Fig. 5** Top 50 Words: “Excellent” tagged hotels.

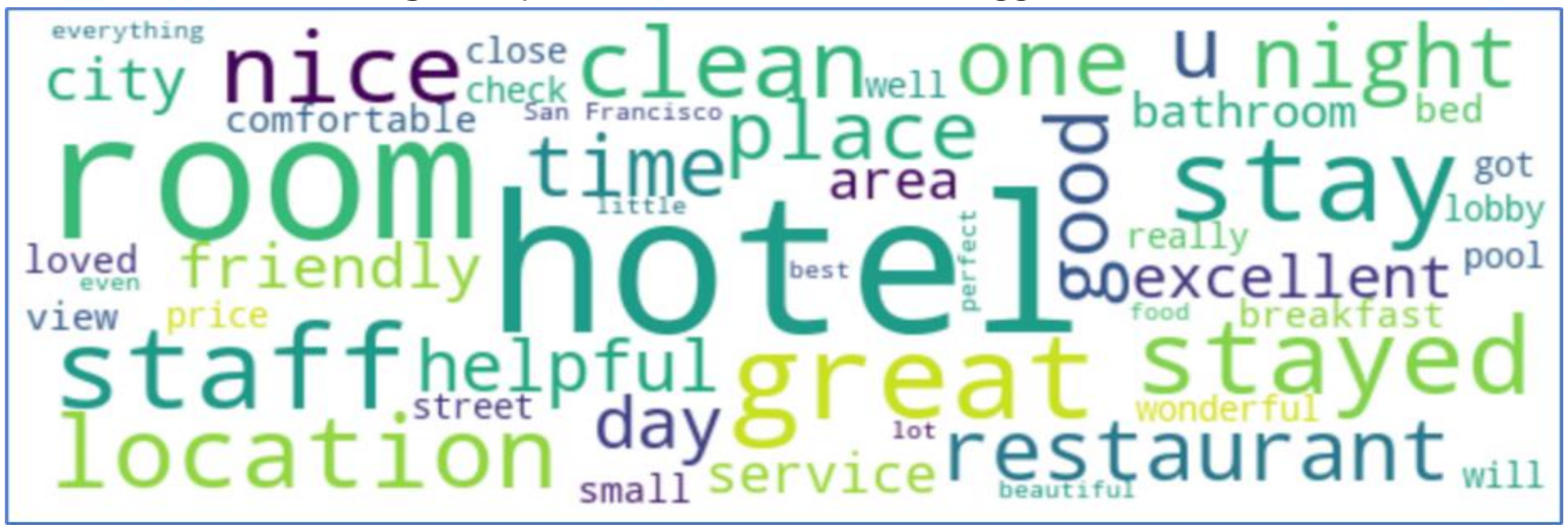

We can identify some words which indicate the hotels as an “excellent” choice, such as “comfortable”, “helpful”, “nice”, “loved”.

## 5. Conclusions and Future Work

In conclusion, a hotel recommendation system can greatly enhance the user experience of booking accommodations by providing personalized recommendations based on user preferences and previous booking history. By utilizing machine learning algorithms and big data analytics, such a system can accurately predict user preferences and suggest hotels that best match their needs. Additionally, the system can also provide real-time updates on availability and pricing to ensure users get the best possible deals. Overall, a hotel recommendation system can streamline the booking process and increase user satisfaction, making it a valuable tool for both customers and hotels.

The proposed approach of using Natural Language Processing (NLP) algorithms, Machine Learning (ML), and Artificial Intelligence (AI) techniques to develop a Decision Support System (DSS) that analyzes customer text reviews is a promising direction for hotel recommendation systems. By leveraging NLP techniques, the system can extract valuable insights from customer reviews, such as sentiment, topics, and key phrases, and use this information to generate personalized recommendations [17-19].

The use of ML and AI algorithms can help to improve the accuracy of these recommendations by taking into account a range of factors, such as the user's budget, location preferences, and past booking history. By incorporating these factors, the DSS can provide tailored recommendations that are more likely to meet the user's needs and preferences. After all, similar works have been done successfully in domains such as cryptocurrency price prediction, patients length of stay and anxiety disorders [20].

Overall, the proposed approach has the potential to greatly enhance the user experience of hotel booking platforms by providing more accurate and personalized recommendations. Additionally, the insights gained from analyzing customer reviews can also be used by hotel managers to improve their services and facilities, leading to increased customer satisfaction and loyalty. There are a few potential drawbacks to take into account, even while hotel recommendation systems based on client internet evaluations can be useful in offering insights and assistance. Online reviews might be biased or manipulated. To sway the ratings, some hotels might publish phony testimonials that are favorable, or rival hotels might do the opposite. Users may be misled by this, which may also reduce the accuracy of the recommendation system. The quality of reviews can differ greatly. It might be difficult to determine the veracity and authenticity of a reviewer's perspective because some reviews may be vague or contain contradictory information. The reliability of the recommendation system may be impacted by this inconsistency. Additionally, hotels with a lot of positive reviews or high ratings may be given priority by recommendation systems, favoring the selections that are more widely accepted. This may reduce the visibility of smaller or more recent businesses that could provide just as excellent or better experiences.

Use hotel recommendation tools that take into account a range of aspects besides user evaluations, such as location, price range, amenities, and personal tastes, to get around some of these constraints. A more comprehensive evaluation of hotels can also be obtained by cross-referencing various sources and using expert comments.

Future research on online review-based hotel recommendation systems has several promising areas. A deeper knowledge of consumer experiences can be achieved by integrating several data modalities, such as text reviews, photos, and videos. More thorough and precise recommendations can be produced by taking the pertinent data from these modalities and integrating it with the others. Also, another future direction due to the advances in Artificial Intelligence, Big Data Analytics and Deep Neural Networks is the provision of Explainability of the recommendations provided by recommender systems to the end user [21-23].

Furthermore, contextual elements, such as the user's present location, the time of day, the weather, or nearby events, can help customize recommendations to the particular context in which users are making their travel selections. The utility and relevance of recommendations may increase as a result. Comparing various recommendation strategies and advancing the subject require the development of comprehensive assessment criteria that capture the efficacy, user happiness, and diversity of hotel recommendations. The goal of these research directions is to improve the personalization, accuracy, and overall user experience of hotel recommendation systems based on online evaluations from guests.